\documentclass[runningheads]{llncs}
\usepackage{graphicx}
\usepackage{mathtools} 
\usepackage{subcaption} 
\usepackage{booktabs}
\usepackage{todonotes}
\usepackage{amssymb}
\renewcommand{\epsilon}{\varepsilon}
\newcommand{\R}{\mathbb{R}}

\newcommand{\sobol}{Sobol\kern-0.15em\'{ } }
\begin{document}
\title{On the Utility of Probing Trajectories for Algorithm-Selection}
%
%
\author{Quentin Renau\inst{1}\orcidID{0000-0002-2487-981X} \and
Emma Hart\inst{1}\orcidID{0000-0002-5405-4413}}

\authorrunning{Q. Renau and E. Hart}

\institute{Edinburgh Napier University, Edinburgh, Scotland, UK 
\email{\{q.renau,e.hart\}@napier.ac.uk}}

\maketitle              
\begin{abstract}
Machine-learning approaches to algorithm-selection typically take data describing an instance as input.
Input data can take the form of features derived from the instance description or fitness landscape, or can be a direct representation of the instance itself, i.e. an image or textual description. Regardless of the choice of input, there is an implicit assumption that instances that are similar will elicit similar performance from algorithm, and that a model is capable of learning this relationship.  We argue that viewing algorithm-selection purely from an instance perspective can be misleading as it fails to account for how an algorithm `views' similarity between instances.
We propose a novel `algorithm-centric' method for describing instances that can be used to train models for algorithm-selection: specifically, we use short \textit{probing trajectories} calculated by applying a solver to an instance for a very short period of time. The approach is demonstrated to be promising, providing comparable or better results to computationally expensive landscape-based feature-based approaches.
Furthermore, projecting the trajectories into a 2-dimensional space illustrates that functions that are similar from an algorithm-perspective do not necessarily correspond to the accepted categorisation of these functions from a human perspective.

\end{abstract}
\section{Introduction}
\label{sec:intro}
We are motivated by the future goal of designing optimisation systems that are capable of learning from past experience. For example, algorithm-selection (AS) methods such as machine-learning based classifiers~\cite{song2023revisit} learn by being trained on results obtained from solving a large set of instances, while transfer-learning methods~\cite{ZhangTransfer,transferL} reuse information extracted from one instance in solving a new instance (e.g. warm-starting with a previous solution).  In both of these scenarios, there is an implicit assumption that if two instances are similar to each other, then they might elicit similar performance from the same solver or that information can be transferred from one instance to another, for example to seed or warm-start an optimiser \cite{kostovska2022per}.

Typical approaches to AS train a machine-learning (ML) model to predict either algorithm-performance or best-solver based on a description of an instance as input to the model in some form. 
There have been many advances made in recent AS literature regarding the choice of model and in defining appropriate model inputs~\cite{kerschke_automated_2019,AlissaSH23}. In particular, a lot of attention has been directed towards defining features to train a model.  Human-designed features can be used, but have the potential disadvantage of being domain-specific and often costly to compute~\cite{KotthoffKHT15}, while it has also been noted that instances that are close in human-designed feature-spaces are not necessarily close in the  performance space of a given algorithm~\cite{SimH22}.  Defining features via Exploratory Landscape Analysis (ELA)~\cite{mersmann_exploratory_2011} has become a popular alternative, particularly in the continuous optimisation domain, creating a feature-vector describing the fitness landscape of an instance. However, there is significant  overhead cost induced by the ELA feature computation: furthermore,  the sample points used to compute features are usually discarded, hence wasting computational budget.  More recent `feature-free' methods avoid calculating features altogether by directly using a description of the instance as input, e.g. using text-based descriptions~\cite{alissa2019algorithm,song2023revisit}  or images~\cite{SeilerPKT22}. However, we argue that all of the above approaches are potentially flawed in that the model is trained on data that takes only an `instance perspective' of the data (through human-designed features, landscape features or a description of the instance data directly): instances are described by features that are calculated independently of the execution data obtained by any algorithm.

Some recent research~\cite{KostovskaJVNWED22,JankovicVKNED22,JankovicED21} has begun to address this, training selection models whose input data includes information derived from running a solver, in addition to (or instead of) using purely instance-centric data.
We continue to push in this direction in proposing a novel method for training a selector that uses \textit{only} time-series information obtained from a  \textit{probing-trajectory}. A  probing trajectory is defined by either the best or current performance of a meta-heuristic solver over its first $n$ function evaluations on an instance, where $n$ is deliberately very short.
We propose that an algorithm that produces similar trajectories on two instances `sees' some commonality between those instances (and vice versa).  Each trajectory (a time-series) can be used as input to an AS classification method, either directly or using time-series features derived from the trajectory. This use of probing-trajectories has the following benefits:
\begin{itemize}
    \item It completely removes the need to either define or calculate features of any type in order to create training data.
    \item The trajectory provides an `algorithm-perspective' of an instance, in contrast to feature-based approaches which only describe the instance (or its associated landscape) in isolation from any solver. We hypothesise that taking the `algorithm perspective' might make it easier for an AS approach to learn as the trajectory is a
    very close proxy to true algorithm behaviour.
    \item The probing-trajectory used to get a prediction from the model can be re-used to warm-start a selected solver, hence saving budget.
\end{itemize}

We evaluate the approach using the BBOB functions~\cite{bbob-functions} as a test-bed.
We show that trajectory-based algorithm-selection can outperform the classical landscape-aware approach in continuous optimisation. In this scenario we also show that for sampling budgets where a landscape-aware approach cannot be applied (i.e., when sampling budgets are too small), trajectory-based algorithm-selection still performs well, making it a good low-budget alternative to ELA features.

The outline of this paper is as follows. Section~\ref{sec:related} gives an overview of the background and related work.
Section~\ref{sec:method} describes the data used, the methods for obtaining probing-trajectories, and describes the experiments conducted in this paper.
Section~\ref{sec:results} describes the results obtained with the probing-trajectories on an algorithm selection task.
Section~\ref{sec:insights} provides insights into trajectory similarities, while Section~\ref{sec:pros} exposes the pros and cons of using the probing-trajectories
Finally, Section~\ref{sec:conclusion} highlights concluding remarks and future work.

\section{Background and Related Work}
\label{sec:related}

The  majority of previous work in algorithm selection is performed using information describing an \textit{instance}, for example extracting features at the instance level; extracting features depicting the landscape of the objective function at hand; using feature-free Deep Learning techniques.  The use of \textit{Instance features} is most common in combinatorial optimisation domains.
They rely intrinsically on the problem domain and are usually manually designed~\cite{HeinsBPSTK23}, differing from one domain to the other, i.e., Travelling Salesperson Problem (TSP) features and Knapsack features cannot be interchanged. However, the main drawback of instance features is that they often do not correlate well with algorithm performance data. For instance, Sim {\em et. al.}~\cite{SimH22} demonstrate in the TSP domain that instances that are close in the feature-space can be very distant in the performance space (i.e., the Euclidean distance between their feature-vectors is small while the distance between the performance of two algorithms on the instances is very large). Furthermore, they can also be computationally expensive~\cite{KotthoffKHT15}.

On the other hand, the use of \textit{landscape-features} is common in numerical black-box optimisation, typically via Exploratory Landscape Analysis (ELA)~\cite{mersmann_exploratory_2011} which calculates landscape features. ELA has grown over the years with a gradual introduction of new features~\cite{munoz_exploratory_2015,kerschke_detecting_2015,DerbelLVAT19}.
Features are numerical values obtained by sampling $m$ points, $x_1, \dots, x_m \in \R^d$ in a $d$-dimensional search space, and computing the associated objective function $f(x_1), \dots, f(x_m)$.
The features are then approximated given the pairs $(x_1,f(x_1)), \dots, (x_n,f(x_m))$.
ELA has been successfully applied to both algorithm configuration~\cite{belkhir_per_2017} and algorithm selection~\cite{kerschke_automated_2019} on benchmark data as well as real-world optimisation problems~\cite{RenauDPSDD22}.
A definition of the most used features and their properties can be found in~\cite{renauPhD}. 
The main drawback of ELA however is the overhead cost induced by the feature computation as the sample points that are usually used to compute features are then discarded.
Other work provides evidence in some domains that ELA features need be used with care~\cite{renauPhD}.

\textit{Feature free techniques} avoid the problem of human-designed features by relying on Deep Learning to extract patterns and be able to perform algorithm selection. Feature free approaches have been successfully applied both in combinatorial optimisation~\cite{AlissaSH23} using the instance definition as input and in continuous optimisation~\cite{SeilerPKT22} using sample points in the search space. While the latter removes the human bias from the design of features, the overhead cost of sampling the search space on top of running the algorithm is still present.
Moreover, Deep Learning approaches reduce the understanding of the instance space and make it difficult to truly understand algorithm behaviour.

All of the approaches just described take an instance-centric view: that is, the input to a selector is independent of the execution of any algorithm. We suggest this is problematic, given that previous work has suggested that there is not necessarily a strong correlation between  the distance of two instances in a feature-space and the distance in the performance space according to a chosen portfolio of solvers \cite{SimH22}. Some recent work has begun to address this, using information derived from running a solver as input to a selector. For example, recent work from Jankovic \textit{et. al.}~\cite{JankovicVKNED22,JankovicED21} proposes  extracting ELA features from the search trajectory of an algorithm.
In~\cite{JankovicED21}, they obtain a trajectory by using half the available budget to run an algorithm ($250$ function evaluations, corresponding to the ELA features recommended budget~\cite{kerschke_low-budget_2016}) and combine this with the state variables of the algorithm to predict the algorithm performance.
Overall, their approach gave encouraging results but was outperformed by classical ELA features computed on the full search space.
In~\cite{JankovicVKNED22}, they use trajectories of $30d$ points to compute ELA features to train a performance predictor in order to choose which algorithm to warm-start.
They successfully compute features during the search of one algorithm to select the appropriate algorithm to switch to finish the run.
This $30d$ points budget is slightly lower than the recommended ELA features budget of $50d$ points.

The work of~\cite{CenikjPDKE23} also obtain algorithm trajectories but then construct time-series based on concatenation of statistics derived from the population and fitness values (mean, standard deviation, minimum, maximum) at each generation to use as input to a classifier that predicts which of the $24$ BBOB function the trajectory belongs to. The length of the trajectories obtained is $900$ points on functions of dimension $d=3$.
This budget is very large as it is $6$ times more than the recommended ELA features budget of $50d=150$.
Their approach successfully outperforms ELA features extracted from algorithm trajectories but is not compared to ELA features extracted from sample points in the full search space.
Another approach combines ELA features with time-series features extracted from state variables of CMA-ES~\cite{KostovskaJVNWED22} to perform a per-run algorithm selection with warm-starting.
The results of this work shows performance on par with ELA features on the per-run algorithm selection task.
Although not specifically concerned  with algorithm-selection, the work of~\cite{PitraRH19} is also worthy of mention in taking an algorithm perspective by utilising information incorporated in CMA-ES \textit{state variables} to train a surrogate model to predict performance.  

We build on the nascent line of work in also proposing a trajectory-based approach to algorithm-selection. Specifically, we attempt to use short trajectories that only consume a small fraction of the available computational budget, and that can be re-used to warm-start an algorithm predicted by a selector. 
Unlike the work described in~\cite{JankovicED21}, we do not compute landscape features from search-trajectories and make use of the probing trajectories both directly and indirectly.

\section{Methods}
\label{sec:method}

\subsection{Data}
\label{sec:data}
We consider the first $5$ instances of the $24$ noiseless Black-Box Optimisation Benchmark (BBOB) functions from the COCO platform~\cite{cocoJournal} as a test-bed.
In BBOB, instances are transformations of the original function such as rotations, translations or scaling.

For each instance, we collect data from running three algorithms: CMA-ES~\cite{HansenO01}, Particle Swarm Optimisation (PSO)~\cite{KennedyPSO95}, and Differential Evolution (DE)~\cite{StornP97}. Each algorithm is run $5$ times per instance. Thus, we have $24\times5\times5=600$ trajectories.
Our data is obtained directly from~\cite{dataDiederick} which records search-trajectories per run. Note that some automated algorithm configuration was performed by the authors before they collected this data and that population sizes are different for each algorithm, see~\cite{Vermetten2022} for further details.

We use data obtained from~\cite{dataEvoStar} on the BBOB suite to calculate ELA features. For each feature, $100$ independent values are available per function instance. The sampling strategy used to sample points is the \sobol low-discrepancy sequence. However, we use only a fraction of the available data, i.e., $10$-dimensional functions, a feature computation budget of $30d$ and the general recommendation for feature computation $50d$~\cite{kerschke_low-budget_2016}.
We tried to compute ELA features at budgets lower than $30d$ but as the budget decreases, Nearest Better Clustering features start to output \emph{Not a Number} values and some Dispersion features are not computed.
We select $10$ cheap features based on their expressiveness and invariance to transformations.
The $10$ features selected are the same as in~\cite{RenauDDD21}.

\subsection{Algorithm Selection Inputs}
\label{sec:input}
Using the data collected in Section~\ref{sec:data}, we create three types of inputs for the algorithm selection procedure: raw-trajectories, features extracted from the time-series formed by the raw-trajectories and ELA features.

\paragraph{Raw Probing-Trajectories}
A probing-trajectory consists of a time-series of values $o$ obtained from the first $n$ iterations of an algorithm, where $o$ is either the current best fitness (coined `current' in the rest of the paper) or the best-so-far fitness values (coined `best' in the rest of the paper). Points are added to the trajectory in the order in which they are sampled\footnote{We evaluate the effect of this choice later in Section \ref{sec:discussion}}.

We evaluate three approaches to using raw probing-trajectories as input to an algorithm-selector: (1) using two different types of probing-trajectory (best-so-far, current); (2) using concatenated probing-trajectories from multiple algorithms as input, i.e., concatenate trajectories from two or more algorithms; (3) input trajectories of different lengths. 
Hence, the input of the algorithm selection procedure is a time-series with its length depending on the number of concatenated algorithms and its number of generations, representing one run on one function instance.

We test the impact of the length of the trajectories on the ability to select the best algorithm using two settings for the number of generations $g \in \{2,7\}$. 
The former enables us to evaluate really short trajectories obtained using only $2$ generations, while the latter results in trajectories using a similar computational budget to that used to obtain ELA features (only one setting described below requires more evaluations than ELA features).
The function evaluations budget depends on the population size of the algorithm according to the dataset used, i.e., $10$ for CMA-ES, $30$ for DE, and $40$ for PSO. For a single trajectory input this results in a budget of ${20,60,80}$ for CMA-ES, DE, PSO respectively at generations = $2$, and ${70,210,280}$ at generations = $7$.  Note that all these budgets are less than the minimum budget of 300 at which it is feasible to compute ELA features. 
For concatenated trajectories, the maximum budget is $560$ when three trajectories are joined ($7 \times (10+30+40)$), therefore comparable to the higher $50d$ budget used to compute ELA features.

\paragraph{Time-series Features}
\label{sec:ts_features}
We extract time-series features from  the probing-trajectories described above as in~\cite{Nobel0B21}, including tuning the process by using a feature selection technique.
The time-series features are extracted on the probing-trajectories using the \emph{tsfresh} Python package (version 0.20.1), while the feature selection is performed using the \emph{Boruta} Python package (version 0.3).

Time-series features are extracted on the same settings as probing-trajectories, i.e., on the `current' or `best' trajectories for a single or concatenated trajectory given $2$ or $7$ generations. The input of the algorithm selection procedure is thus a vector of features where the dimension depends on the number of features extracted or selected, representing one run on one function instance.

\paragraph{ELA Features}
\label{sec:ela}
As mentioned above (in Section~\ref{sec:data}), $100$ $10-$dimensional feature vectors are available for each BBOB instance.
These features are computed with $30d = 300$ and $50d = 500$ sample points.
In order to have a fair comparison between ELA and probing-trajectories, we randomly sample $5$ feature vectors out of $100$ to match the number of available trajectories by instances, i.e, $5$ runs by instances.

The input of the algorithm selection procedure is a $10-$dimensional vector of features, representing one function instance.

\subsection{Algorithm Selection Methods}
\label{sec:classif}
The algorithm selection is treated as a classification task, i.e, given an input representing an instance, the output is the algorithm to use on that particular instance.
To train the classification models, both the trajectories and feature-vectors are labelled with the best performing algorithm, defined as the one having the best median target value after $100{,}000$ function evaluations.
Note that no single algorithm outperforms the others on all $24$ functions: CMA-ES is the best performing algorithm for $11$ functions, DE for $7$, and PSO for $6$.

In Section~\ref{sec:input}, we described two types of input (trajectories and feature vectors), hence two types of classifiers are used depending on the data type.

\paragraph{Classification with Features}
Features inputs can be derived from calculating either ELA features or time-series features extracted from the probing-trajectories.
As used in~\cite{BelkhirDSS17,RenauDPSDD22}, we train a default Random Forests~\cite{Breiman01} from the \emph{scikit-learn} package~\cite{scikit-learn} (version 1.1.3). 
Separate models are trained using different input type, i.e., models using ELA features and time-series features are different.

\paragraph{Classification with Raw Trajectories}
The raw probing trajectories form a time-series. For this type of input, a specialised time-series classifier is used, specifically the default Rotation Forests~\cite{RodriguezKA06} from the \emph{sktime} package (version 0.16.1).
This choice of classifier is motivated by its closeness to the classifier used for features while accounting for the data being time-series.
We train classifiers using single or concatenated trajectories to predict which algorithm of the portfolio to use, e.g., a classifier trained with `CMA-ES' trajectories can predict which of the three algorithms from the portfolio to use.

\paragraph{Validation Procedure}
Both type of inputs have the same validation procedure.
As performed in~\cite{KostovskaJVNWED22}, we perform a \emph{leave-one-instance-out (LOIO) cross-validation} and we compute \emph{the overall accuracy}.
Classifiers are trained using data from all runs of the $24$ functions on all except one instance.
The data of the left out instance is used as the validation set, i.e, all the runs from the $24$ functions on the ID of the left out instance.

Overall, $480$ inputs are used to train the model while the remaining $120$ inputs are used for validation.

\section{Results}
\label{sec:results}
In this section, we first present the results obtained with single trajectories (Section~\ref{sec:single}), followed by results obtained from using input  to a classifier that is formed by concatenating trajectories from multiple algorithms (Section~\ref{sec:concat}).  In each case, we compare the classification results  using three types of input: the raw-trajectory; a feature-vector derived from the time-series; an ELA feature-vector. We discuss the results in Section~\ref{sec:discussion}.

\subsection{ELA Features vs Single Trajectories}
\label{sec:single}
In this section, we compare algorithm-selection performed using ELA features as input to trajectory-based inputs obtained from running a single algorithm. For the latter, we experiment with using inputs from (1) the raw trajectory data and (2) time-series features derived from the trajectory. We train three separate classifiers, using trajectories from each of the algorithms in turn\footnote{A classifier trained only on e.g. CMA-ES trajectories can predict any of the three solvers, etc.}.

Algorithm trajectories are computed for $2$ and $7$ generations. 
As explained previously, the minimum budget is $20$ function evaluations for CMA-ES ($2$ * population size $10$) and the largest is $280$ function evaluations for PSO ($7$ * population size $40$).
As mentioned in Section~\ref{sec:data}, ELA features are extracted using $30d = 300$ and $50d = 500$ points to provide fair comparisons.

We observe that models trained with best-so-far trajectories outperforms models trained with current trajectories with $2$ generations and vice versa with $7$ generations.
Due to space limitations, we will only present best-so-far trajectories for $2$ generations and current trajectories for $7$ generations.
The other plots can be found in the supplementary materials.

Figure~\ref{fig:single_loio} compares the classification accuracy of the classifiers trained using probing-trajectories, time series features extracted from three trajectories and time series feature selection.

Classifiers trained using only $2$ generations (using a maximum budget of 80 evaluations)  exhibit a poor accuracy (Figure~\ref{fig:single_low_loio}).
Classifiers trained on ELA features  perform better but recall that these require a minimum budget of $300$ evaluations, almost $4$ times the budget for the trajectories. Even with this additional budget, the median accuracy reached is relatively poor: $90\%$ and $90.8\%$ for $300$ and $500$ function evaluations respectively.
This is not unexpected and consistent with previous literature, given that is known that ELA features are not all invariant to function transformations~\cite{renauPhD,SkvorcEK22}.
One of the trajectory-based classifier models exceeds the performance  of the ELA trained classifier at a budget of $300$ samples and matches the ELA trained classifier with $500$ samples, demonstrating $90.8\%$ median accuracy: PSO with raw probing-trajectories. In this context, the use of a PSO probing-trajectory is clearly an asset as it requires less than a sixth of ELA features budget to achieve the same level of performance.

When the number of generations increases to $7$ (Figure~\ref{fig:single_large_loio}), all but one of the classifiers trained on input obtained from a trajectory outperform both the ELA trained classifiers (budget $300, 500$).  
A considerable increase in performance accuracy is obtained in most cases. Once again, the PSO trajectories reach the best performances with a peak for raw probing-trajectories at a median accuracy of $100\%$.
In this case, almost perfect classification is achieved with substantially less function evaluations than needed to compute ELA features.

PSO trajectories may be more informative than other algorithms trajectories because PSO has the largest population size and evaluates more points.
As a comparison, CMA-ES has the smallest population size and is in most cases the algorithm providing the worse accuracies.
Hence, the information contained in the trajectories may be a matter of the number of generations and the number of points evaluated.

\begin{figure}
\centering
\begin{subfigure}{.45\textwidth}
  \centering
  \includegraphics[width=\linewidth]{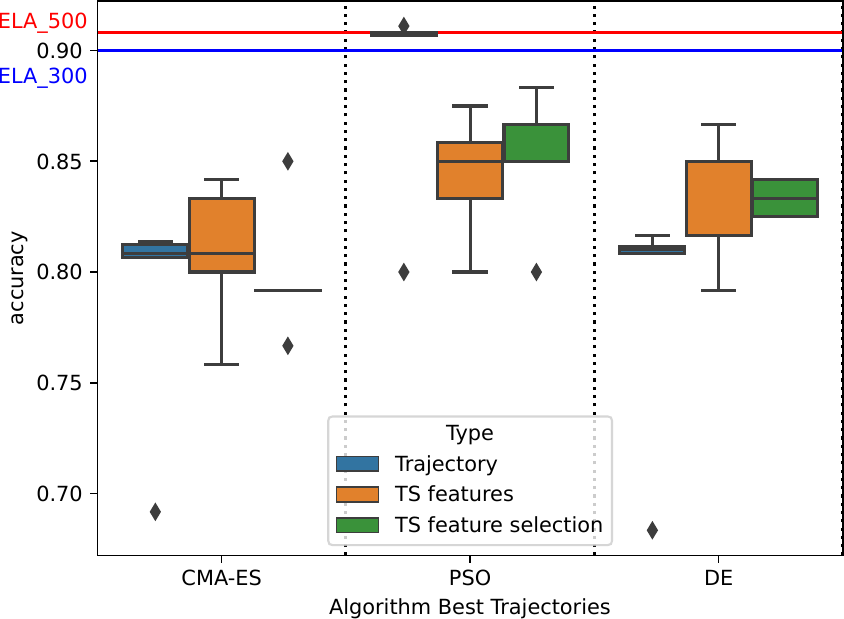}
  \caption{Best trajectory for $2$ generations}
  \label{fig:single_low_loio}
\end{subfigure}
\begin{subfigure}{.45\textwidth}
  \centering
  \includegraphics[width=\linewidth]{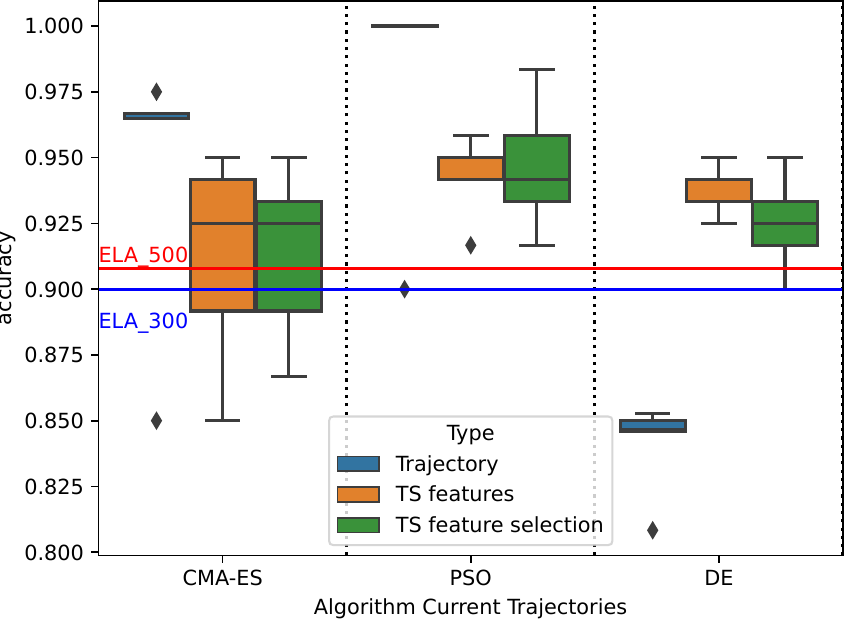}
  \caption{Current trajectory for $7$ generations}
  \label{fig:single_large_loio}
 \end{subfigure}
 \caption{Accuracy of classification on the LOIO cross-validation for best-so-far and current probing-trajectories, time series features and time series feature selection for $2$ and $7$ generations. Median ELA feature accuracy is represented by lines for $300$ and $500$ function evaluations.}
 \label{fig:single_loio}
\end{figure}

\subsection{ELA Features vs Multiple Trajectories}
\label{sec:concat}

In this section, we compare algorithm-selection performed with ELA and time series features with classifiers using a concatenation of trajectories as input (i.e., the trajectories obtained from more than one algorithm are joined and used as input to the classifier to predict the best algorithm).
By concatenating trajectories, we train four models named: $C-P$ (for the concatenation of CMA-ES and PSO trajectories), $C-D$ (for CMA-ES and DE), $D-P$ (for DE and PSO) and, $ALL$  (for the concatenation of the three algorithms).
These models are also trained for $2$ or $7$ generations.

As in Section~\ref{sec:single}, we observe that models trained with best-so-far trajectories outperforms models trained with current trajectories with $2$ generations and vice versa with $7$ generations.
Due to space limitations, we will only present best-so-far trajectories for $2$ generations and current trajectories for $7$ generations.
The other plots can be found in the supplementary materials.

Figure~\ref{fig:concat_loio} compares the classification accuracy of the concatenation of trajectories on the LOIO cross-validation.
As observed previously in Section~\ref{sec:single}, classifiers trained using the raw trajectories outperform those trained using the time-series features and ELA feature selection.

At $2$ generations (Figure~\ref{fig:concat_low_loio}), classifiers trained on raw trajectories outperform those trained on ELA features except in a single case ($C-D$).
The classifiers trained with time-series features are all outperformed by classifiers trained on ELA features when using this low budget of $2$ generations.

When the number of generations increases to $7$ (Figure~\ref{fig:concat_large_loio}), the picture changes dramatically. Classification accuracies increase and all trajectory-based classifiers outperform classifiers using ELA features. Recall that the maximum budget for the trajectory based input here is $560$ ($ALL$) and the minimum $280$ ($C-P)$ compared to the ELA budgets of ($300,500$). While accuracies of time series-based classifiers are comparable, the median accuracy of raw trajectories increases to $100\%$ (with one run around $90\%$ for all classifiers).

\begin{figure}
\centering
\begin{subfigure}{.45\textwidth}
  \centering
  \includegraphics[width=\linewidth]{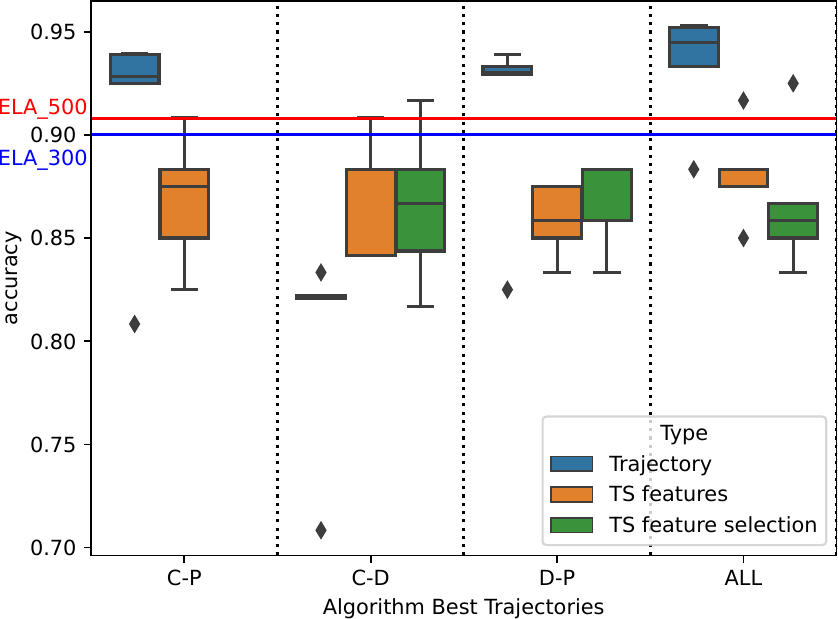}
  \caption{Best trajectory for $2$ generations}
  \label{fig:concat_low_loio}
\end{subfigure}
\begin{subfigure}{.45\textwidth}
  \centering
  \includegraphics[width=\linewidth]{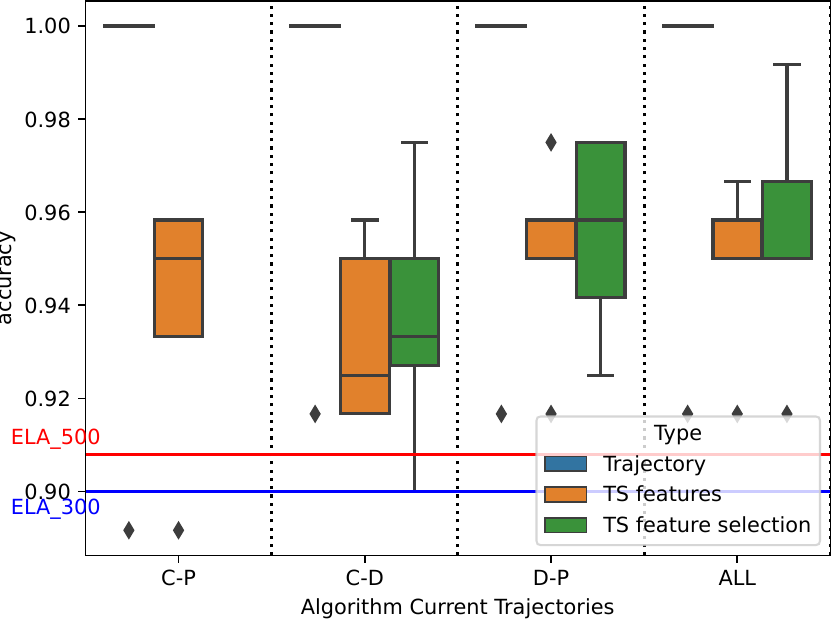}
  \caption{Current trajectory for $7$ generations}
  \label{fig:concat_large_loio}
 \end{subfigure}
 \caption{Accuracy of classification on the LOIO cross-validation for best-so-far and current probing-trajectories, time series features and time series feature selection for $2$ and $7$ generations. Median ELA feature accuracy is represented by lines for $300$ and $500$ function evaluations.}
 \label{fig:concat_loio}
\end{figure}

\subsection{Additional Insights into Trajectory Performance}
\label{sec:discussion}

\paragraph{Randomness in the `current' trajectory}
As noted in Section \ref{sec:input}, the `current' trajectory is obtained by adding each point sampled by an algorithm  to a trajectory in the order in which they are evaluated. However, it should be clear that for a generational algorithm, the order that solutions are evaluated in is irrelevant.
Therefore to satisfy ourselves that the particular ordering used had no influence of the results, we perform an additional experiment in which the points within each trajectory are randomly shuffled within a generation to obtain a new trajectory.
We consider trajectories of $7$ generations for the three algorithms in the portfolio. 
This process is repeated $5$ times.

We repeat previous experiments  using the raw trajectory, and time-series features as input to a classifier.
We performed a Kolmogorov-Smirnov statistical test between the initial run and the $5$ shuffled trajectories. The p-values for the five tests performed are between $0.84$ and $1$.
These values do not enable the null hypothesis to be rejected at a confidence level of $0.05$, thus we conclude that the order the points are evaluated in the `current' trajectories does not impact the classification accuracy.

\paragraph{Order in the $ALL$ trajectory}
In the same manner, the concatenated $ALL$ trajectories used in the experiments used a fixed ordering to obtain the concatenation, i.e., CMA-ES, PSO, and DE ($C-P-D$). 
Therefore, we also investigate the impact of the order used to construct the $ALL$ trajectory to determine if it has an impact on the classification accuracy. We evaluate the classification accuracy of the six possible combinations for the $ALL$ trajectory.
We use the raw trajectory, compute time series features and perform feature selection on the LOIO cross-validation using the `current' trajectory on $7$ generations as before.
We performed a Kolmogorov-Smirnov statistical test between the combination used in previous sections ($C-P-D$) and all other combinations.
The p-values for the five tests performed are between $0.87$ and $1$.
These values do not enable the null hypothesis to be rejected at a confidence level of $0.05$, thus we can conclude that the order used to create the $ALL$ trajectory does not impact the classification accuracy.

\begin{figure}
\centering
\includegraphics[width=0.7\linewidth]{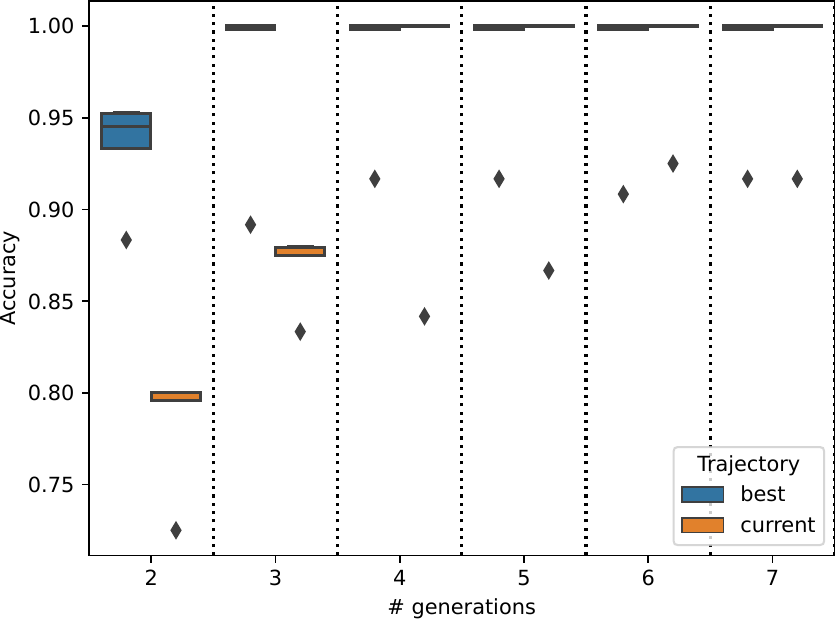}
 \caption{Boxplot of accuracy of classification of the `best' and `current' trajectory-based classifiers on $ALL$ trajectories from $2$ to $7$ generations on LOIO cross-validations.}
 \label{fig:best_vs_current}
\end{figure}

\paragraph{Best-so-far vs current trajectories}
We mention in Section~\ref{sec:single} and Section~\ref{sec:concat}, that the `current' trajectory outperforms the `best' for larger number of generation and vice versa.
In Figure~\ref{fig:best_vs_current}, we display accuracy of classification of the $ALL$ trajectories for different generations from $2$ to $7$.

We observe that the crossing point between the `best' and `current' trajectories happens with $4$ generations.
With fewer generations, the `best' trajectory outperform the `current' until they reach similar accuracies with $4$ generations.
Interestingly, we observe that only $3$ generations are necessary for the $ALL_{best}$ to reach a median accuracy of $99.8\%$ (with one run at $89.2\%$) which is $9\%$ better than ELA features for half their computation budget.

An explanation of the difference in performance of the two trajectories resides in the fact that the `best' trajectories can be seen as elitist (i.e, keeping only the best value) and `current' trajectories can be seen as non elitist (i.e., accepting lower fitness values).
It has been seen that elitist algorithms are outperforming non elitist ones for small budgets and vice versa when the budget increases (an example of this behavior can be found in~\cite{RenauDPSDD22}).
Hence the `best' trajectory may contain more useful information when the budget is low whereas more generations may be needed for the `current' trajectory to reach the same level of information.

\section{Insights into Trajectory Similarity}
\label{sec:insights}

To understand  whether similarity between trajectories correlates with similarity in terms of solver performance, we project the time-series obtained into 2 dimensions using UMAP from the \emph{umap-learn} Python library (version 0.5.3) with default parameters in an unsupervised setting.

\begin{figure}[h]
\centering
\begin{subfigure}{.45\textwidth}
  \centering
  \includegraphics[width=\linewidth]{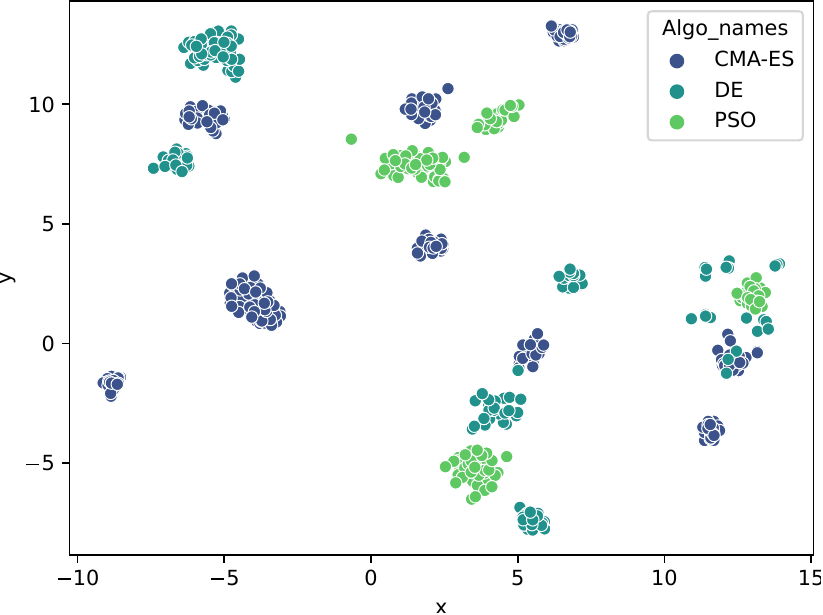}
  \caption{Algorithms}
  \label{fig:umap_traj}
\end{subfigure}
\begin{subfigure}{.45\textwidth}
  \centering
  \includegraphics[width=\linewidth]{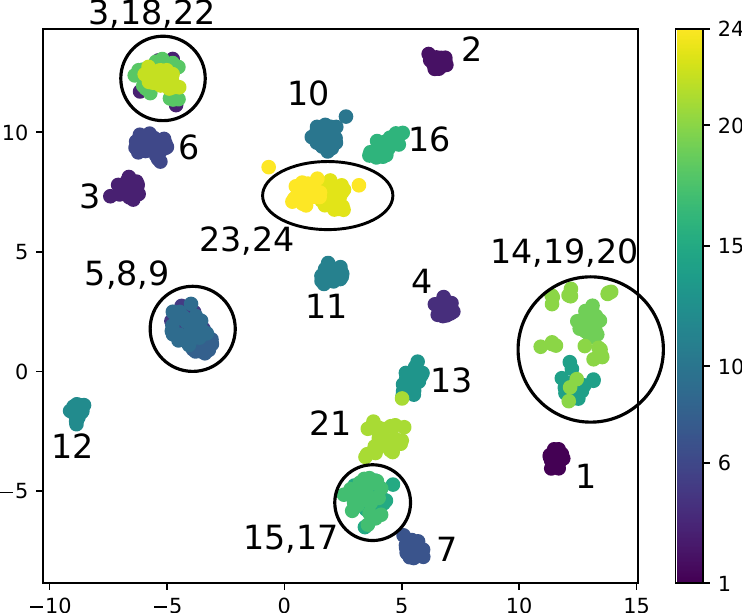}
  \caption{BBOB functions}
  \label{fig:umap_fid}
 \end{subfigure}
 \caption{UMAP 2d projection of the $ALL_{current}$ time series of runs on function instances for $7$ generations for each algorithm and BBOB functions ($560$ function evaluations).}
 \label{fig:umap}
\end{figure}

Figure~\ref{fig:umap} presents projections for the $ALL_{current}$ time series using a budget of $560$ function evaluations to create the trajectories, i.e. $7$ generations.
Each point represents a trajectory for one run on an instance.
Colour labels in Figure~\ref{fig:umap_traj} represent the best algorithm for the function the trajectory was obtained from, i.e., the label used during classification for a given trajectory as in Section~\ref{sec:classif}.

For a given algorithm, multiple clusters can be seen throughout the space.
This may imply that for a given algorithm, distinct types of trajectories lead to the same choice of algorithm. However, we also observe some clusters of points where several  algorithms are grouped in the same cluster.
This may affect the accuracy of classification: in particular the figure suggests it may induce errors between CMA-ES and DE, and between PSO and DE.

The number of clusters is lower than the number of BBOB functions, implying that from the algorithm perspective, different functions may be similar. In order to verify the similarity of functions from the algorithm perspective, we coloured the instance runs by function.
The result is shown in Figure~\ref{fig:umap_fid}. 
From the algorithm perspective, there are clear groups of functions won by the same algorithm. Trajectories from only one function (F3) are found in different groups.  F3 belongs to two groups that are close in the space and hence may actually be one larger group.

From a trajectory perspective, ten functions have their own group (F1, F2, F4, F6, F10, F11, F12, F13, F16, and F21) while the others are distributed in groups of at least two functions.
The largest groups are composed of three functions that are similar from a trajectory point of view.
For two of these groups, all functions are won by the same algorithm: DE for F3, F18, and F22 and CMA-ES for F5, F8, and F9.
Interestingly, these functions belong to different categories (BBOB functions are divided into five categories representing function properties) in the BBOB test-bed~\cite{cocoJournal}: three different categories for F3, F18, and F22 and two for 5, F8, and F9.
Recall that these categories are human-designed: however, the results imply that the properties of the functions that are observable by a human do not completely reflect the algorithm perspective. 
This reinforces the point made in Section~\ref{sec:intro} that determining instance similarity from an algorithm perspective might prove useful in future attempts to build optimisation systems that are capable of learning across instance sets.

\section{Pros and Cons of Trajectories}
\label{sec:pros}
The results in the previous sections show that trajectory-based algorithm selection is a good low-budget counterpart to classical approaches that tend to use ELA features. At very low budget ($2$ generations, $80$ evaluations), a classifier trained on raw PSO trajectories performs on par with a classifier based on ELA features obtained using $500$ samples.  Three classifiers trained with concatenated trajectories $(C-P, D-P, ALL)$ obtained from $2$ generations with total budgets of $100, 140,160$ respectively also outperform the ELA classifiers.  Using a larger number of generations ($7$), $8$ out of $9$ classifiers trained on single raw trajectories or information derived from the trajectory outperforms the ELA based classifier using $500$ samples.  Using concatenated trajectories from $7$ generations, all classifiers outperform the ELA classifiers, with the classifiers trained on concatenated raw, `current' trajectories obtaining $100\%$ median accuracy (with a maximum budget of $560$ evaluations). 

Given that computation of ELA features  usually requires approximately $30d$ samples and that the computation is known to fail for some features at budgets less than this, using trajectories to train selectors is a promising way forward, providing reasonable results even at very low budget. In addition, the trajectory obtained to provide input to the selector can also be re-used in some cases: for example, if a classifier trained on `CMA-ES' trajectories predicts CMA-ES for an instance, then the run can simply be continued from where it was stopped. Furthermore, if using  the $ALL$ trajectory which contains partial runs of all algorithms in the portfolio, then the budget is `virtually' reduced by a third as any selected algorithm can simply continue its run from the already computed trajectory.

Comparing the results of the $ALL$ classifiers that use three concatenated trajectories with results from trajectories formed from pairs of classifiers raises the question of how the method might scale with the number of algorithms in the portfolio. For example, for a portfolio of $N$ solvers, it is possible that only a subset of trajectories is sufficient to train a classifier. This necessitates consideration of how to select the most appropriate combination of trajectories, i.e., $M < N$.  As the fixed evaluation budget $b$ must be divided  between $N$ trajectories (such that each trajectory is allocated $o=b/N$ function evaluations), then the influence of settings of $o$ and $N$ for a fixed value $b$ should also be studied further. Finally, further investigation of how the approach scales as the $N$ increases is also required.

\section{Conclusion}
\label{sec:conclusion}
The paper address the issue of algorithm-selection in a continuous optimisation, specifically in a setting whether there is a fixed  budget of evaluations that must be shared between any computation required to derive input to an algorithm-selector and in running the selected algorithm. The goal is to find a form of input to a classifier that requires minimal computational budget, delivers  high accuracy, and views similarity from an algorithm perspective. We hypothesised that short probing-trajectories obtained by running an algorithm for a small number of generations could be used to train a classifier, with the added benefit that the run used to obtain the trajectory could simply be continued for the chosen algorithm.

We demonstrated that a time-series classifier trained on raw trajectories and classifiers trained on features extracted from the trajectories can outperform classifiers trained on ELA features at considerably lower budget, in the best case using six times fewer sample points.
Moreover, unlike using ELA features where sample points used for feature computation are often discarded, points from trajectories can be re-used if the algorithm that a trajectory belongs to is selected or points can be re-used for warm-starting another optimisation algorithm.

Obvious next steps include testing the approach on a more complicated task with larger algorithm portfolios as well as on different data-sets. A more thorough evaluation of state-of-the-art time-series classifiers (and tuning of their hyper-parameters) is also likely to improve results. 
Another key question lies in the scaling of the $ALL$ concatenation of trajectories approach: does the number of trajectories needed depend linearly on the size of the portfolio or can judicious selection of specific trajectories suffice?  Another interesting direction for future work is to use evaluate the use of the approach in combinatorial optimisation domains where there is much less work in defining appropriate ELA features and where calculation of domain-specific features is often expensive (particularly in TSP). Furthermore, in some combinatorial domains, hand-designed features have also been criticised for not correlating well with performance~\cite{SimH22}.
A trajectory-based approach could therefore be a promising avenue for research.

\section*{Acknowledgements}
The authors are funded by the EPSRC `Keep-Learning' project:  EP/V026534/1 and EP/V027182/1

\bibliographystyle{splncs04}
\bibliography{references}

\end{document}